\documentclass[a4paper,twoside]{article}

\usepackage{amsmath,amssymb,amsfonts,algorithmic,graphicx,textcomp,xcolor,hyperref,soul}
\usepackage{multirow}
\usepackage{balance}
\usepackage{epsfig}
\usepackage{subcaption}
\usepackage{calc}
\usepackage{amstext}
\usepackage{amsthm}
\usepackage{multicol}
\usepackage{pslatex}
\usepackage{apalike}
\usepackage{float}
\usepackage{stfloats}
\usepackage{SCITEPRESS}     
\begin{document}

\title{A Case Study and Qualitative Analysis of Simple Cross-Lingual Opinion Mining}


\author{
    \authorname{
        Gerhard Hagerer\sup{1}\orcidAuthor{0000-0002-2292-0399}, 
        Wing Sheung Leung\sup{1},
        Qiaoxi Liu\sup{1},
        Hannah Danner\sup{2}\orcidAuthor{0000-0001-8387-0818} and 
        Georg Groh\sup{1}\orcidAuthor{0000-0002-5942-2297}
    }
    \affiliation{
        \sup{1}Social Computing Research Group, Department of Informatics, Technical University of Munich, Germany
    }
    \affiliation{
        \sup{2}Chair of Marketing and Consumer Research, TUM School of Management, Technical University of Munich, Germany
    }
    \email{~}
}



\abstract{User-generated content from social media is produced in many languages, making it technically challenging to compare the discussed themes from one domain across different cultures and regions. It is relevant for domains in a globalized world, such as market research, where people from two nations and markets might have different requirements for a product. We propose a simple, modern, and effective method for building a single topic model with sentiment analysis capable of covering multiple languages simultanteously, based on a pre-trained state-of-the-art deep neural network for natural language understanding. To demonstrate its feasibility, we apply the model to newspaper articles and user comments of a specific domain, i.e., organic food products and related consumption behavior. The themes match across languages. Additionally, we obtain an high proportion of stable and domain-relevant topics, a meaningful relation between topics and their respective textual contents, and an interpretable representation for social media documents. Marketing can potentially benefit from our method, since it provides an easy-to-use means of addressing specific customer interests from different market regions around the globe. For reproducibility, we provide the code, data, and results of our study\footnote{\url{https://github.com/apairofbigwings/cross-lingual-opinion-mining}}.}

\onecolumn \maketitle \normalsize \setcounter{footnote}{0} \vfill

\section{INTRODUCTION}

Topic modeling on social media texts is difficult, since lack of data as well as spelling and grammatical errors can make the approach unfeasible. Dealing with multiple languages at the same time adds more complexity to the problem which oftentimes makes the approach unusable for domain experts. Thus, we propose a cross-lingually pre-trained deep neural network as a black box with very little textual pre-processing necessary before embedding the texts and forming their clustering and topic distributions. 

For our method, we leverage current research regarding multi-lingual topic modeling, see Section \ref{sec:relatied_work}. We provide an extensive description of a simple method to support domain experts from specific social media domains in its application in Section \ref{sec:topic_modeling_method}. 
We qualitatively demonstrate our topic model, its feasibility, and its cross-lingual semantic characteristics on English and German newspaper and social media texts in Section \ref{sec:cross_lingual_topic_coherence}. We aim at inspiring pragmatic ideas to explore the potential for comparative, inter-cultural market research and agenda setting studies. Unsolved problems and future potential are given in Section \ref{sec:conclusion}.

\begin{figure*}[t]
\centerline{\includegraphics[width=0.95\linewidth]{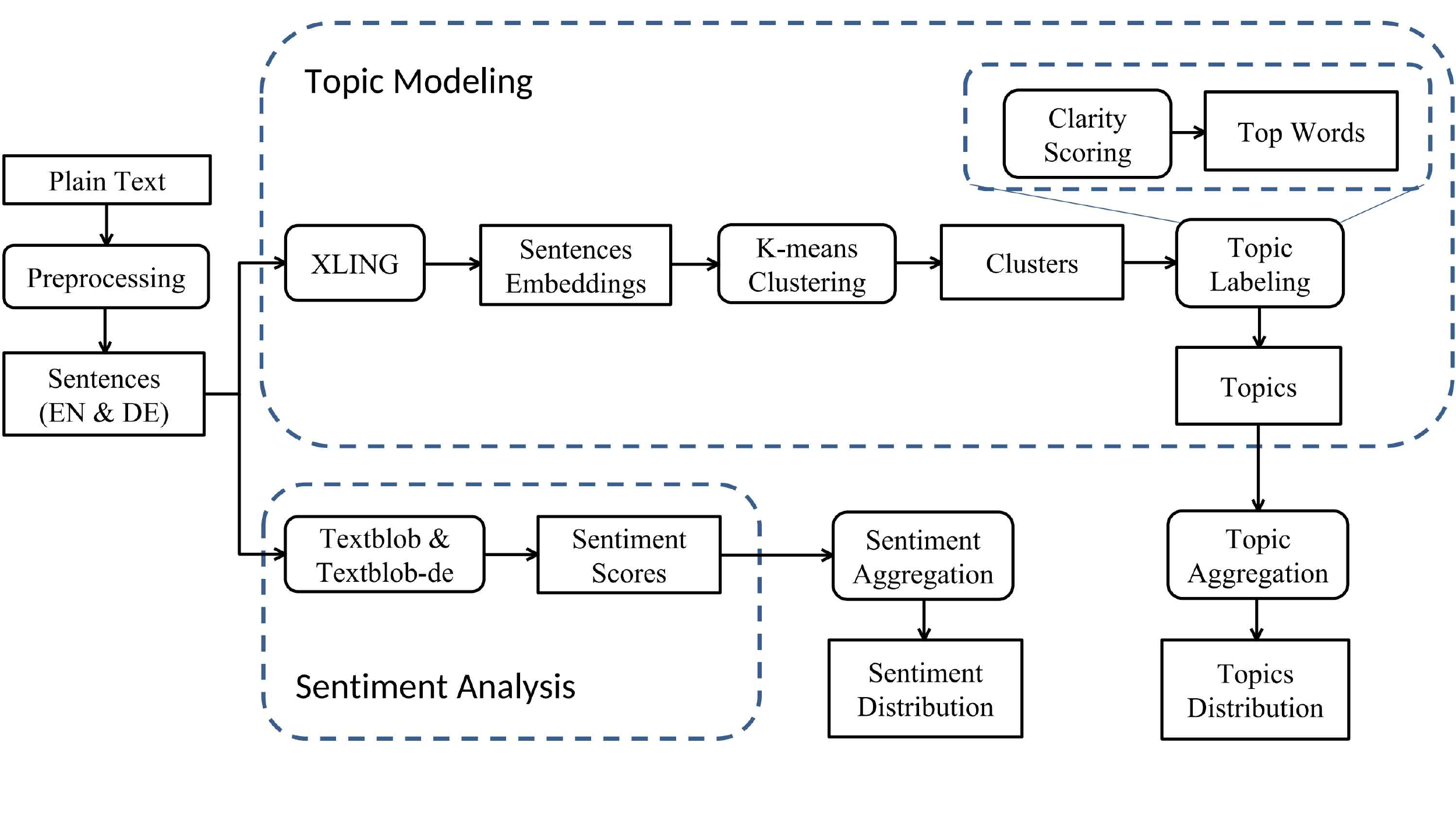}}
\vspace{-0.6cm}
\caption{Plain text is first tokenized into sentences and passed to topic modeling and sentiment analysis. Topic modeling involves (1) converting sentences of both languages into embeddings with XLING, (2) clustering all embeddings with K-means and (3) deriving a topic label of each cluster. Sentiment analysis is performed using Textblob. Topic and sentiment scores are aggregated for the analysis.}
\label{fig:workflow}
\vspace{-0.2cm}
\end{figure*}

\vspace{-0.2cm}

\section{RELATED WORK} \label{sec:relatied_work}

\vspace{-0.1cm}

Topic modeling is meant to learn thematic structure from text corpora. With probabilistic topic modeling methods, such as latent semantic indexing (LSI) \cite{LSI} or latent Dirichlet allocation (LDA) \cite{LDA}, researchers try to extend the capabilities of topic modeling for application from a single language to multiple languages. Using multi-lingual dictionaries and translated corpora is an intuitive way to tackle cross-lingual topic modeling problems \cite{21c,21d}. Further ~examples exist with either ~dictionaries or ~translation \\ text collections \cite{21a,21b,21e}. However, this puts dependence on the availability of dictionary or good quality of translations. Significant manual labor and verification are required to prevent deteriorating noise.


Recently, methods converting words to vectors according to their semantics are widely adopted \cite{word2vec}. Several studies showed text embeddings improve topic coherence \cite{22e,22ei}. Regarding multi-linguality, embeddings in word level and sentence level enable text in different languages to be projected to the same vector space \cite{universalSentenceEncoder} such that semantically similar texts are clustered together independently of their languages. This favors studies on multi-lingual topic modeling without relying on dictionaries and translation \cite{22a,22b}. Although providing highly coherent topics, a recreation of word spaces is required when new text corpora are introduced. In our scenario, these limitations are not present.

Regarding the application of topic modeling, various social media corpora are studied by domain experts \cite{23b,23c}. They covered different domains, such as politics, marketing, and public health. Regarding media agenda setting, \cite{23a} studied on how much degree a Russian newspaper related to economic downturn. They also \textit{"introduced embedding-based methods for cross-lingually projecting English frame to Russian"} based on Media Frames Corpus. 
In contrast, we propose a straightforward topic modeling method without fine-tuning but only clustering necessary on a social media corpus. This enables further investigation on media agenda setting cross-lingually and cross-culturally.

\section{TOPIC MODELING METHOD} \label{sec:topic_modeling_method}

Figure \ref{fig:workflow} shows the overall workflow of our topic modeling approach. We aim to conduct simple, cross-lingual topic modeling on user-generated content with no translation, dictionary, and parallel corpus required for aligning the semantic meanings across languages. Our approach solely depends on clustering sentence embeddings for topic modeling. Ready-made sentence representations simplify the approach, since these suppress too frequent, meaningless, and unimportant words automatically without the need to model that part explicitly \cite{kim2017bag}.


\subsection{Preprocessing}
The raw texts of articles and comments are first tokenized into sentences with Natural Language Toolkit (NLTK). Then, URLs, specially for those enclosed with HTML \verb|<a>| tag, are replaced with string 'url'. After that, sentences with character length smaller than 15 are omitted to minimize noise, since they appear inscrutable and they are only 6.6\% out of all sentences which is a small portion. After preprocessing, there are 127,464 English sentences and 200,627 German sentences, i.e., total 328,091.

\begin{figure}[t]
\centerline{\includegraphics[width=\columnwidth]{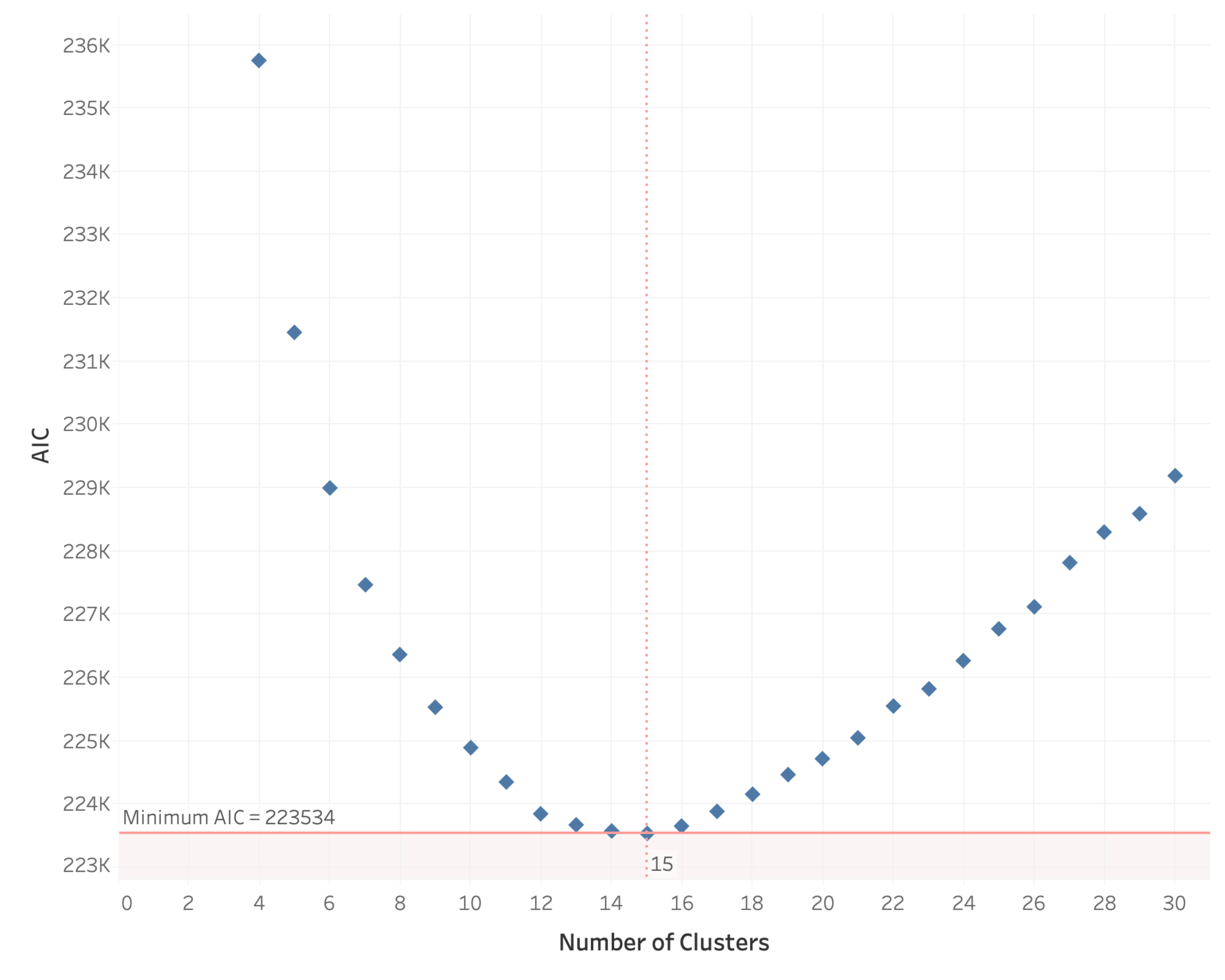}}
\caption{AIC plot indicates $k$ = 15 is the global minimum.}
\label{fig:AIC}
\vspace{0.3cm}
\end{figure}

\subsection{Cross-Lingual Embeddings}

In the following paragraph, we provide an explanation of the pre-trained XLING model, which we use for the present work, based on the words of the authors \cite{XLING-en-de}. XLING calculates \textit{"sentence embeddings that map text written in different languages, but with similar meanings, to nearby embedding space representations"}. Similarity is calculated mathematically as dot product between two sentence embeddings. In order to train the model, the authors \textit{"employ four unique task types for each language pair in order to learn a function g"}, i.e., the eventual sentence-to-vector model. The architecture is based on a Transformer neural network \cite{vaswani2017attention} tailored for modeling multiple languages at once. The tasks on which the model is eventually trained are \textit{"(i) conversational response prediction, (ii) quick thought, (iii) a natural language inference, and (iv) translation ranking as the bridge task"}. The data for training \textit{"is composed of Reddit, Wikipedia, Stanford Natural Language Inference (SNLI), and web mined translation pairs"}.

\subsection{Sentence Clustering}
K-means clustering algorithm is implemented on both English and German sentence embeddings at the same time. Since XLING provides semantically aligned sentence embeddings of both languages, this joint clustering step helps to establish one topic model for two disjunct datasets irrespective of their language. Clustering is established for a varying number of clusters$k$, ranging from 1 to 30. Elbow method is first used for choosing the optimal $k$ but the inertia (sum of squared distances of samples to their closest cluster center) of increasing $k$ decreases rapidly at the beginning and then gently without a significant elbow point. Therefore, Akaike Information Criterion (AIC) is adopted and $k$ = 15 is chosen as optimal value as it is the global minimum, see Figure \ref{fig:AIC}. 
In Section \ref{sec:semantic_relation}, further discussion on topic coherence is conducted proving the fact that $k$ = 15 resulted semantically coherent topics.


\subsection{Topic Labeling} \label{sec:clarity_scoring}
To be able to derive a meaningful topic label for each sentence cluster, the respective top words of each cluster are required. In order to get the top word list, the clarity score is adopted \cite{clarityscoring}.
According to \cite{Angelidis2018SummarizingOA}, it ranks terms with respect to their importance of each cluster $c$ and language $l$, such that

\begin{equation}
\mathrm{score}_{l,c}(w) = t_{l,c}(w) \log_{2}\frac{t_{l,c}(w)}{t_l(w)},
\end{equation}

where $t_{l,c}(w)$ and $t_l(w)$ are the l1-normalized tf-idf scores of the word $w$ in the sentences within cluster $c$ and in all sentences, respectively, for a certain language.

Additionally, stopword removal from the top word lists is also a concern when calculating the clarity score. Generally, stopwords are the most frequent words in the documents and sometimes they are too dominant such that they interfere with the result from clarity scoring. Thus, we remove domain-specific high frequency words for each language from corresponding topic top word lists.

Topics are labeled manually based on the English and German top word lists. The results are shown in Table \ref{tab:topwords} and will be discussed further in Section \ref{sec:semantic_relation} evaluating topic coherence across languages.

\begin{table*}[t]
\caption{Top words for all meaningfull topics with $k$ = 15 of English and German data}
\centering
\begin{center}
\begin{tabular}{|p{0.13\linewidth}|p{0.36\linewidth}|p{0.42\linewidth}|}
\hline
\textbf{Topic} & \textbf{English top words} & \textbf{German top words} \\
\hline
Environment & pesticide, plant, soil, use, crop, fertilize, pesticide, garden, herbicide, grow & pflanze, pestizid, dunger, boden, gulle, garten, anbau, gemuse, tomate, feld \\
\hline
Retailers 
& store, whole, shop, groceries, supermarket, local, market, amazon, price, online, discount 
& aldi, supermarkt, lidl, kauf, laden, lebensmittel, cent, einkauf, wochenmarkt %
\\ \hline
GMO ~~~~~~~~~~~~~~\& organic & gmo, label, gmos, monsanto, product, certificate, usda, genetic, product & produkt, bioprodukt, lebensmittel, gesund, konventionell, biodiesel, herstellung, enthalt, monsanto, pestizid \\
\hline
Food products \& taste 
& taste, milk, sugar, cook, eat, fresh, flavor, fruit, potato, sweet 
& kase, schmeckt, gurke, essen, analogkase, schmeckt, tomate, milch, geschmack, kochen 
\\ \hline
Food safety & chemical, cancer, body, acid, effect, cause, toxic, toxin, glyphosat, disease & dioxin, gift, grenzwert, ddt, menge, giftig, toxisch, substanz, chemisch, antibiotika \\
\hline
Research 
& science, study, scientific, research, gene, scientist, genetic, human, stanford, nature %
& gentechnik, natur, mensch, wissenschaft, lebenserwartung, genetisch, studie, menschlich, planet \\
\hline
Health ~~~~~~~~~~~~\& nutrition & eat, diet, healthy, nutritious, health, fat, calory, obesity, junk & lebensmittel, essen, ernahrung, gesund, nahrungsmittel, lebensmittel, nahrung, fett, billig \\ 
\hline
Politics ~~~~~~~~~~~~~~\& policy & govern, public, politic, corporate, regulation, law, obama, vote & politik, skandal, verantwortung, bundestag, schaltet, bestraft, strafe, kontrolle, kriminell \\ 
\hline
Animals ~~~~~~~~~~~~\& meat & meat, chicken, anim, cow, beef, egg, fed, raise, pig, grass & tier, fleisch, eier, huhn, schwein, futter, kuh, verunsichert, vergiftet, deutsch \\
\hline
Farming & farm, farmer, agriculture, land, sustain, crop, yield, acre, grow, local & landwirtschaft, landwirt, bau, flache, okologisch, nachhaltig, konventionell, landbau, produktion, ertrag \\ 
\hline
Prices \& profit 
& price, consume, market, company, profit, product, cost, amazon, money 
& verbrauch, preis, produkt, billig, qualitat, kunde, kauf, geld, unternehmen, kosten \\
\hline
\end{tabular}
\label{tab:topwords}
\end{center}
\end{table*}

\begin{figure*}[t]
\centerline{\includegraphics[width=\linewidth]{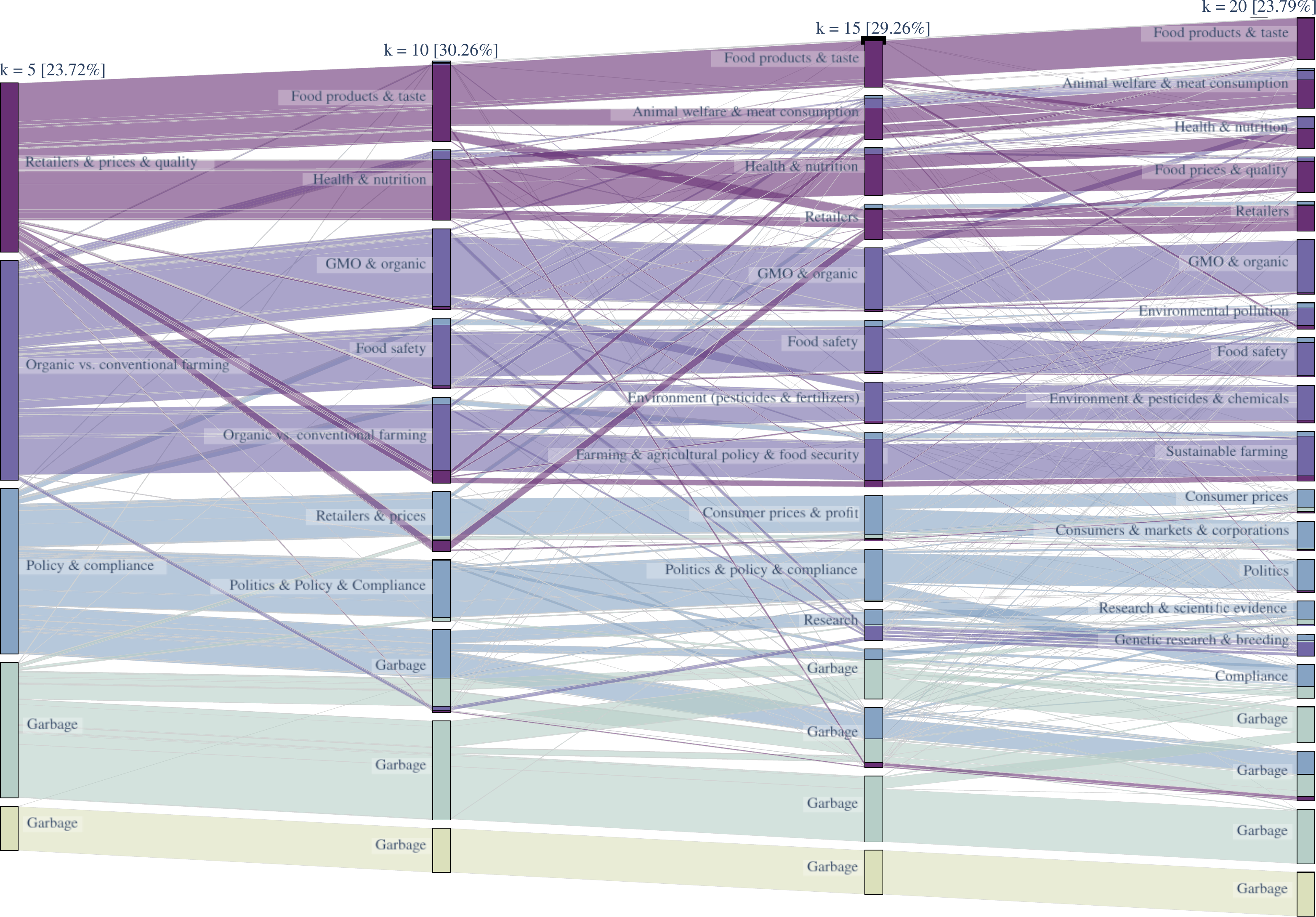}}
\caption{Topic distributions with increasing number of topics $k$. The percentage is the amount of sentences in garbage topics.}
\vspace{-0.3cm}
\label{fig:sankey}
\end{figure*}

\subsection{Sentiment Analysis}

In addition to topic modeling, we conduct sentiment analysis to investigate the feasibility and meaning of cross-lingual topic-related sentiments in articles and respective comment sections. We make use of Textblob\footnote{https://github.com/sloria/textblob} and Textblob-de\footnote{https://github.com/markuskiller/textblob-de} to assign each of the English and German pre-processed sentences a polarity score. The polarity assignment is first proposed by \cite{sentimentSubjectivity} and reimplemented by \cite{textblob}. Since the subjectivity assignment is not well-developed in Textblob-de, we filter out sentences with polarity equals to 0 for both English and German sentences in order to derive comparable results.

\subsection{Topic and Sentiment Distributions}
After assigning a labeled cluster, i.e., a topic, and a sentiment score to each sentence of the corpus, we derive the corresponding distributions.



For topic distributions, all sentences from a document are counted per topic. The distribution is then normalized to be comparable. 
For sentiment distributions, all sentences from a document are grouped per topic. 
Topic-wise sentiment distribution is derived based on the sentence-wise polarity scores and the respective median and quartiles.
A document in that regard is either an article or all of its comments, i.e., its comment section.

\begin{figure*}[t]
\centerline{\includegraphics[width=\linewidth]{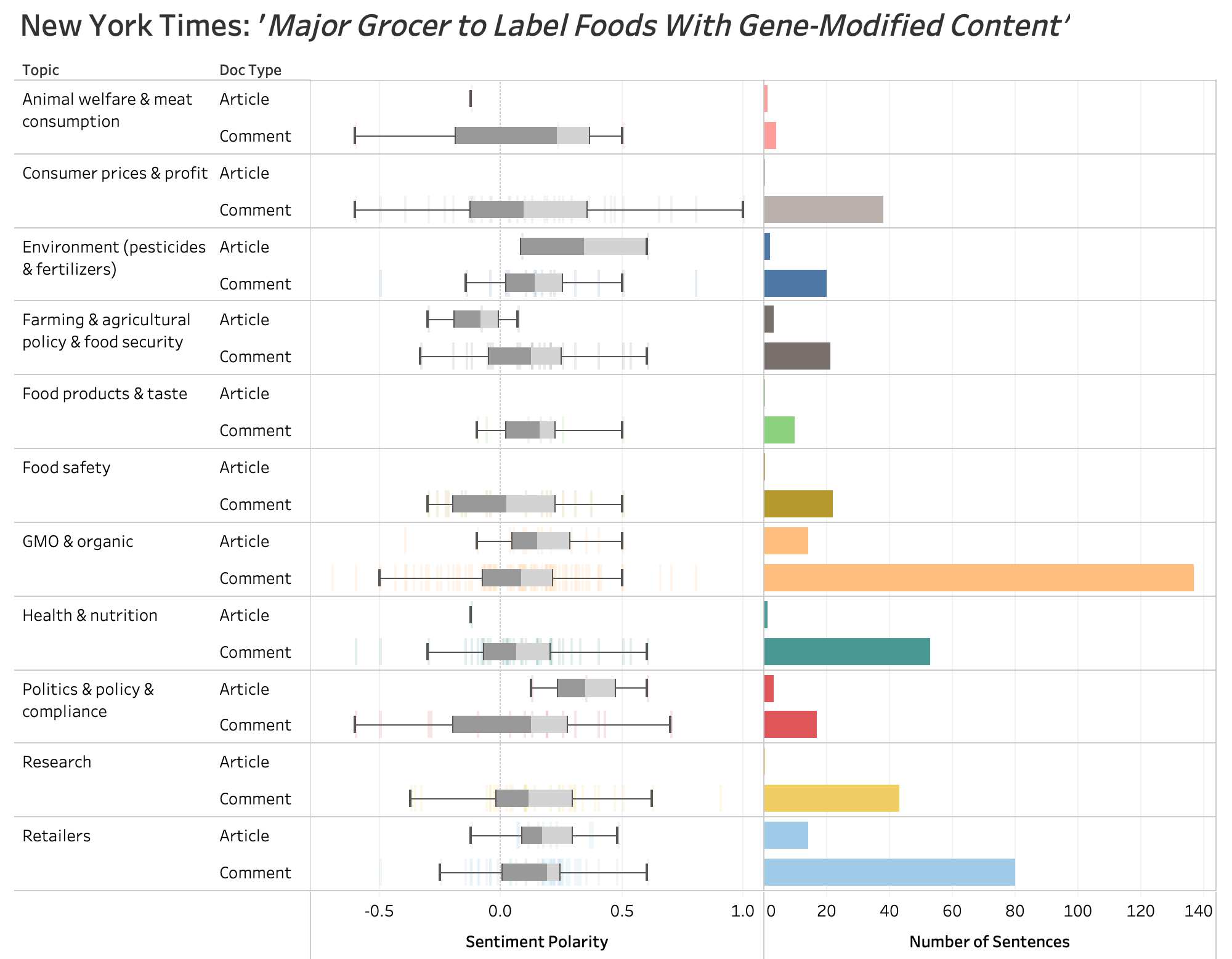}}
\caption{Topic and sentiment distribution for \textit{Grocer}.}
\label{fig:sentiment_distribution_en}
\vspace{-0.2cm}
\end{figure*}

\section{TOPIC COHERENCE} \label{sec:cross_lingual_topic_coherence}

In this section, we evaluate the feasibility and semantic coherence of our cross-lingual topic modeling qualitatively. Instead of providing quantitative coherence scores, we aim at a detailed, qualitative analysis of textual examples. We depict representative sentences and words of each topic in subsection \ref{sec:semantic_relation}. We investigate to what extent these are semantically coherent, also across languages. We expose the ratio of coherent and incoherent topics and how it develops with increasing number of topics in in subsection \ref{sec:usable-clusters}. Eventually, we show the distribution of topics in selected newspaper articles and their respective comment sections to relate the discussed content with our actual topic model on English and German texts.

\subsection{Data}

The collection of the data used in this study is described in another publication \cite{danner2021newsmedia} as follows.
For the analysis we downloaded \textit{"news articles and reader comments of two major news outlets representative of the German and the United States (US) context"}, i.e., \textit{spiegel.de} and \textit{nytimes.com}. The creation dates of the texts are \textit{"spanning from January 2007 to February 2020"}. \textit{"Articles and related comments on the issue of organic food were identified using the search terms \textit{organic food} and \textit{organic farming} and the German equivalents}. For topic modeling, we utilized \textit{"534 articles and 41,320 comments from the US for the years 2007 to 2020, and 568 articles and 63,379 comments from Germany for the years 2007 to 2017 and the year 2020"}.

\begin{figure*}[t]
\centerline{\includegraphics[width=\linewidth]{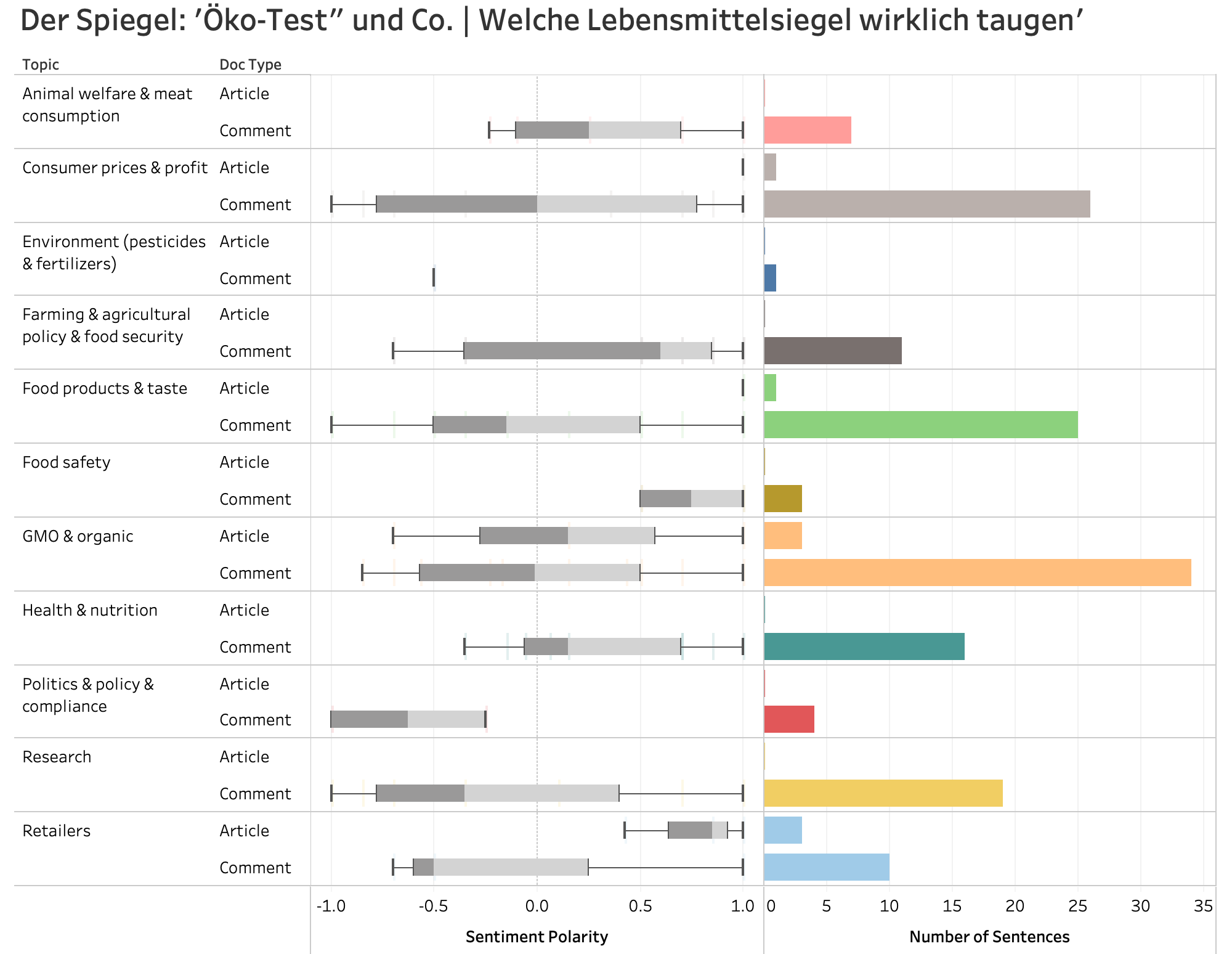}}
\caption{Topic and sentiment distribution for \textit{\"Oko-Test}.}
\label{fig:sentiment_distribution_de}
\vspace{-0.2cm}
\end{figure*}

\subsection{Multi-Linguality of Topics} \label{sec:semantic_relation}

In this section, we evaluate semantic coherence of our cross-lingual topic modeling by depicting the representative sentences and words for each topic and showing the semantic relation. Table \ref{tab:topwords} shows the first 10 English and German words having the highest clarity scores (see Section \ref{sec:clarity_scoring}) in each cluster for $k$ = 15. Table \ref{tab:top_sentences} shows the first 3 English and German sentences whose embeddings have the largest cosine similarity to their corresponding cluster centroids. Both top words and top sentences indicate that the clusters are grouped reasonably in terms of semantics. For example, this is the case for the topic \textit{Environment (pesticides \& fertilizers)} which is indeed related to use of pesticides in planting. Even though this also appears to be the case for the sentences in \textit{GMO \& organic} on the first glance, those are actually about organic food and how aspects such as GMO and pesticides relates to the food itself. This and the other representative top words and sentences indicate that clustering on cross-lingual sentence embeddings yield semantically coherent topics.

According to our analysis, top sentences from garbage clusters are always short in length with slightly more than 15 characters. Together with top words (Table \ref{tab:topwords}), these hardly contribute to the organic food domain and corresponding entities.
Thus, it is feasible in our case to ignore them.

\subsection{Amount of Meaningful Topics} \label{sec:usable-clusters}
Besides providing coherent cross-lingual topics, our method performs well to distinguish usable from unusable topics, and it provides a constantly high number of relevant topics independently of the number of overall topics. Figure \ref{fig:sankey} is a Sankey diagram showing the flow of topic assignments for all English and German sentences with increasing number of clusters $k$. Topic modeling is performed for each $k$ with all pre-processed sentences independently. It can be seen that more specific topics descend from general but related topics as indicated by the colors. 

For instance, \textit{GMO \& organic}, \textit{Food safety}, \textit{Environment (pesticides \& fertilizers)} and \textit{Farming \& agricultural policy \& food security} for $k$ = 15 are derived from \textit{Organic vs. conventional farming} for $k = 5$. \textit{Organic vs. conventional farming} in $k = 5$ generally focuses on advantages brought by organic farming when comparing to conventional farming, such as reducing persistent toxin chemicals from entering to environment, food, and bodies; thus, bio-products are recommended. For $k = 15$, the children topics are more specific. For example, \textit{GMO \& organic} shows the aims for having organic food, i.e., avoidance of GMO and poisoning with pesticides. Moreover, \textit{Food safety} in $k = 15$ is further split into \textit{Food safety} and \textit{Environmental pollution}. 

To see how the topics relate to their actual sentences, we try to observe the top sentences of each topic, i.e., those sentences of which the embeddings are closest to the centroid. Both English and German sentences are similar and share strong semantic similarity. The \textit{food safety} topic focuses on the toxicity issue of dioxin and other chemicals towards consumers. \textit{Environmental pollution}, which is further splitted from it, for $k = 20$ indeed tells contamination of water resources by chemicals. This shows that fine-grained topics and the way they develop with increasing $k$ have a meaningful relation to ancestor and sister topics.

Sentences without contribution to the organic food domain always remain in garbage clusters in a way that the proportion of usable and unusable clusters does not fluctuate. Thus, the topic model maintains its coherence independent to the number of topics and the despite the fact that k-means is not deterministic in its clustering. This property is helpful, since the number of topics can be chosen as high as necessary to provide a sufficient level of detail for the domain of interest. Moreover, this highlights the meaningfulness and robustness of the given sentence representations being able to separate noise from informative data in an unsupervised fashion.

\subsection{Validation of Opinion Distributions}

In this section, two real text examples are given to evaluate our method qualitatively. The first one is an article from New York Times, titled '\textit{Major Grocer to Label Foods With Gene-Modified Content}'\footnote{\url{https://www.nytimes.com/2013/03/09/business/grocery-chain-to-require-labels-for-genetically-modified-food.html}}, hereafter referred to as \textit{Grocer}. It reported that the first retailer in the United States announced to label all of its genetically modified food sold in its stores. Advocating and opposing stakeholders stated their arguments regarding different aspects. The second example is from Der Spiegel, titled '\textit{"\"Oko-Test" und Co. -- Welche Lebensmittelsiegel wirklich taugen}'\footnote{\url{https://tinyurl.com/spiegel-lebensmittelsiegel}}, below denoted as \textit{\"Oko-Test}. It reported that number of food claims, certifications, and seals in Germany were growing as organic labeling was a good promotional strategy indicating high food quality. However, consumers knew little about the details even when the tests for each label were transparent and well-documented. Based on these two summaries, it would be expected that topics related to supermarkets, retailers, and GMO labels are shown to be present in those articles. The \textit{Grocer} article, however, expresses concerns about the consumption of genetically modified food, whereas \textit{\"Oko-Test} discusses organic food labeling issues from various point of views, among others fair trading and organic fishing.

\paragraph{Topic Distribution} 

Figures \ref{fig:sentiment_distribution_en} and \ref{fig:sentiment_distribution_de} show the distribution of topics in the overall article sentences. It can be seen that the two topics \textit{Retailers} and \textit{GMO \& organic} are mentioned the most in both articles, supporting our hypothesis. The comment section of the \textit{Grocer} article corresponds to the article itself such that most of its sentences also talk about \textit{GMO \& organic} and the second most for \textit{Retailers}. However, the commenters of the \textit{\"Oko-Test} articles commented more about \textit{GMO \& organic} followed with \textit{Consumer prices \& profit} and \textit{Food products \& taste}. Even though the dominating topics in German differ between article and comments, it can be stated that the topic distribution overall still refers to the actual topics of the given texts and domains. At the same time, differences in the distribution not only between article and comments but also between languages and thus cultures are directly visible, providing a means for clear comparability in several respects.


\paragraph{Sentiment Distribution} 

Figures \ref{fig:sentiment_distribution_en} and \ref{fig:sentiment_distribution_de} also show the sentiment distribution. Generally, the sentiment of the \textit{Grocer} article spreads out less than that of the \textit{\"Oko-Test} article. It is observed that, in topic \textit{GMO \& organic}, comments score sentiment polarity ranging between $0.50$ to $-0.70$ in \textit{Grocer} and between $1$ and $-0.85$ in \textit{\"Oko-Test}. This means sentences from \textit{Grocer} show weaker sentiment compared to those from \textit{{\"Oko}-Test}. The actual texts indicate that sentiment on our German data indeed has more variance than on English. Thus, the proposed multi-lingual sentiment analysis, Textblob and Textblob-de, appears to represent the data adequately in the given use case. However, it cannot be excluded that the sentiment distribution could be affected given the fact that two different but methodically similar frameworks are used. Different biases and variances could be caused by different models which have differences in the sentiment dictionary size and the subjectivity of human-assigned sentiment scores based on different cultures. Further studies should examine this problem for more robust, domain-independent multi-lingual sentiment prediction.

\section{CONCLUSION} \label{sec:conclusion}

This case study shows that our technically simple approach successfully generates an high proportion of relevant and coherent topics for our domain, i.e., organic food products and related consumption behavior based on English and German social media texts. Moreover, the topics display the text contents correctly and support a domain expert in the content analysis of social media texts written in multiple languages. 

However, the presented paper did not provide quantitative measurements of topic coherences and comparisons with the state-of-the-art. For mono-language topic modeling, it would be LDA \cite{LDA}; for advanced cross-lingual topic modeling, it could be attention-based aspect extraction \cite{he-etal-2017-unsupervised} utilizing aligned multi-lingual word vectors \cite{conneau2017word}. Several multi-lingual datasets would need to be included for a representative comparison. Since pre-trained models trained on external data are used for the proposed method, it might be relevant for coherence score calculation to include intrinsic coherence scoring methods based on train test splits, such as, UMass coherence score \cite{mimno-etal-2011-optimizing}, and explore extrinsic methods calculated on external validation corpora, e.g., Wikipedia \cite{roder2015exploring}.

Regarding multi-lingual sentiment analysis, the difference in the sentiment analysis frameworks for different languages must be considered. For example, since two independent but similar sentiment analysis models are applied for English and German, the sentiment distribution could be affected. Therefore, future studies on developing and evaluating comparable sentiment models should be conducted.


\bibliographystyle{apalike}
{\small
\bibliography{main}
}

\newpage

\begin{table*}[t]
\section*{APPENDIX}
{\color{white}\_}
\tiny
~ ~ ~
\caption{Top sentences of meaningful topics from the whole dataset for $k$ = 15 in English and German}
~ ~ ~
\label{tab:top_sentences}
\centering
\begin{center}
\setlength\tabcolsep{1pt}
\begin{tabular}{|p{0.05\linewidth}|p{0.93\linewidth}|}
\hline 
\textbf{Topic} & \textbf{Top 3 sentences for English and German} \\ 
\hline 
\multirow{2}{\linewidth}{Environ- ment (pesti- cides, fertilizers)} & Usually, the plant which uses conventional farming will produce the residue of the pesticides. 
 -- Some pesticides used in conventional farming, however, may reduce the level of resveratrol in plants. 
 -- Also, there is the question of naturally occurring pesticides produced by the plant itself. \\ \cline{2-2} 
 & Viele Biopflanzen werden zwar nicht mit Pestiziden behandelt, Ihnen wird jedoch sehr viel mehr Wachstumsfl\"ache zugestanden. 
 -- Nicht nur Biobauern benutzen G\"ulle, und Herbizide und Pestizide werden vor allem in der konventionellen Landwirtschaft eingesetzt. 
 -- Zum einen bauen sich die Pestizide und Herbizide relativ schnell ab, nicht zu verwechseln mit \"Uberd\"ungung durch G\"ulle oder Belastung mit Schwermetallen. \\ \hline
\multirow{2}{\linewidth}{Retailers} & Whole Foods also sells a lot of high quality grocery items that aren’t available elsewhere in a lot of places. 
 -- Whole Foods executives, however, say their supermarkets can be high quality, organic and natural but also inexpensive. 
 -- Larger competitors like Safeway and Kroger have vastly expanded their store-brand offerings of natural and organic products, and they are often cheaper than those at Whole Foods. \\ \cline{2-2}
 & Auch bei ALDI und CO lassen sich hochwertige Lebensmittel erwerben. 
 -- Wobei ich feststellen muss, dass andere Superm\"arkte - zumindest die, die ich frequentiere - auch Wert darauf legen, das gewisse Produkte aus der Region stammen, auch wenn sie konventionell hergestellt wurden. 
 -- Der Trend zur Feinkost beschert dem Handel vor allem in den Großst\"adten steigende Ums\"atze, wo Bio-L\"aden hip sind und die kaufkr\"aftigen Kunden beim Einkaufen nicht auf jeden Cent schauen. \\ \hline
\multirow{2}{\linewidth}{GMO \& organic} & In a nutshell, though, organic means the product meets a number of requirements, such as no GMOs, no non-organic pesticides, etc. 
 -- When consumers buy organic, they are guaranteed little more than food that is (in theory at least) produced without synthetic chemicals or G.M.O.’s (genetically modified organisms), and with some attention (again, in theory) to the health of the soil. 
 -- Organic food includes products that are grown without the use of synthetic fertilisers, sludge, irradiation, GMOs, or drugs, which already shows how much better it is for health. \\ \cline{2-2}
 & Jeder weiß doch, daß der Vorteil von bio nicht in der erh\"ohten Aufnahme von N\"ahrstoffen gegen\"uber konventionellen Produkten liegt, sondern in der Vermeidung, sich mit Pestiziden zu vergiften. 
 -- Bio-Lebensmittel genießen einen guten Ruf, weil sie wesentlich weniger Schadstoffe enthalten als konventionell hergestellte Lebensmittel. 
 -- Bioprodukte sind kaum ges\"under als konventionelle Lebensmittel \\
 \hline
 \multirow{2}{\linewidth}{Food products \& taste} & It tastes totally different from your normal vegetables. 
 -- It can also be mixed in with the other foods (milk and fruit, oats/rice cooked in milk). 
 -- The increased flavor is the result of the food containing more micronutrients. \\ \cline{2-2}
 & Die meisten frischen Zutaten m\"ussen etwas aufbereitet werden, damit sie gegessen werden k\"onnen. 
 -- Es braucht wirklich nur Mehl, Wasser und etwas Salz, keinerlei andere Inhaltsstoffe, Punkt. 
 -- Nichts geht \"uber frisch zubereitete Speisen aus gesunden Zutaten. \\ \hline
 \multirow{2}{\linewidth}{Food safety} & Dioxins are extremely toxic chemicals, and their bioaccumulation in the food chain may potentially lead to dangerous levels of exposure. 
 -- Many of the toxins found in non-organic foods are toxins that have a cumulative effect on our bodies. 
 -- As proved by various researchers, these chemicals have harmful effects not only on the consumers but also on the environment and farmers. \\ \cline{2-2}
 & Tatsache ist die Dosen um die es bei Nahrungsmittelkontaminationen durch Dioxine geht sind derart gering dass sie auf keinem Wege zu eine signifikanten Gesundheitsgefahr f\"uhren. 
 -- Dioxine sind UBIQUIT\"AR und entstehen in nicht unerheblichen Mengen durch nat\"urliche Vorg\"ange. 
 -- Es bedarf erheblicher Dosen um gef\"ahrliche Effekte von Dioxin und Lindan nachzweisen. \\ \hline
 \multirow{2}{\linewidth}{Research} & Science is not always applied in benign ways, even when we know as much - growth hormones and indiscriminate use of antibiotics in livestock, for example. 
 -- The bottom line is that genetically modified organisms have not been proven in any way to be safe, and most of the studies are actually leaning the other direction, which is why many of the world’s countries have banned genetically engineered foods. 
 -- However, evolution and adaptation, especially to unknown and unnatural substances, takes many generations for humans to achieve. \\ \cline{2-2}
 & Zu wenig verstehen die Wissenschaftler noch von \"okologischen und evolution\"aren Prozessen. 
 -- Hinzu kommt dass die Natur st\"andig neue genetisch ver\"anderte Organismen hervorbringt. 
 -- Es geht nicht um die Behauptung von Gentechnik sondern um die Behauptung ihrer Resultate. \\ \hline
 \multirow{2}{\linewidth}{Health \& nutrition} & While the government wants us to eat healthy, it is very true that organic foods are outrageously priced for the small amount of food we recieve. 
 -- The health issue with foods lies in our collective wisdom that insists on making foods as cheap as possible. 
 -- Of course there are health benefits to eating “organic” food. \\ \cline{2-2}
 & F\"ur die Bev\"olkerungsschichten die auf g\"unstige Lebensmittel angewiesen sind es komplizierter sich gesund zu ern\"ahren. 
 -- Im Vordergrund unserer Lebensmittelwirtschaft steht eben der Profit und nicht die gesunde Ern\"ahrung. 
 -- Es ist sowieso viel ges\"under Lebensmittel zu essen, die einen m\"oglichst geringen Verarbeitungsgrad aufweisen. \\ \hline
 \multirow{2}{\linewidth}{Politics \& policy \& com- pliance} 
 & There are more that "government regulators" involved. 
 -- We full well know that the industry in all of its glory takes precedence over the concerns or welfare of the people of this country. 
 -- This is all being decided in PRIVATE, There is no involvement by the political or judicial processes that normally make laws in this country. \\ \cline{2-2}
 & Lobbyismus m\"usste als Straftatbestand angesehen werden und \"ahnlich schwerwiegend behandelt werden wie Landesverrat. 
 -- Das ist das Ergebnis der Lobbyarbeit und unsere Volksvertreter verabschieden solche Strafrahmen nicht versehentlich, sondern ganz bewußt. 
 -- Das und \"ahnliches, \"andert nichts an der kriminellen Energie der durch die Politik und Gesetzgebung Vorschub geleistet wird. \\ \hline
 \multirow{2}{\linewidth}{Animal welfare \& meat con- sumption} & Organically raised animals used for meat must be given organic feed and be free of steroids, growth hormone and antibiotics. 
 -- When it comes to meat, again organic is the better option as animals are often treated cruelly and inhumanely to increase production. 
 -- Manure produced by organically raised animals wreaks less havoc on the environment, but the meat may still wreak havoc on arteries. \\ \cline{2-2}
 & Diejenigen die noch Fleisch essen nehmen Bio weil diese Tiere etwas weniger gequ\"alt werden als wie im konventionellen Bereich. 
 -- Im Falle von Fleisch geht das nicht anders, als dass man Tiere qu\"alt und dazu mit Dingen f\"uttert, die man kaum noch als Futter bezeichnen kann. 
 -- Rinder fressen nat\"urlicherweise kein Zucht-Getreide wegen des hohen St\"arke und Fettgehalts. \\ \hline
 \multirow{2}{\linewidth}{Farming \& agricultural policy \& food security} & Regardless of capacity the modern crops still need more and more land to feed the more and more people, even if the inefficiencies and failures of corporate agriculture are overcome. 
 -- The main problem in organic farming is the availability of adequate organic sources of nutrients (crop residues, composts, manures) to supply crops with all the required nutrients and to maintain soil health. 
 -- Without this, the organic farming industry can’t sustain economically as most of the Organic food that is produced is bought by the big packaged food companies. \\ \cline{2-2}
 & Daß der D\"unger f\"ur Bio-\"Acker n\"amlich von Nutztieren hergestellt wird und mehr Bedarf daran auch mehr Nutztiere zur Folge hat, geh\"ort nicht zu den Notwendigkeiten, mit denen die Bio-Lobby hausieren geht. 
 -- Wegen der F\"orderung von Biogasanlagen und dem dadurch entstandenen Bedarf etwa an Mais sei Ackerland inzwischen vielerorts so teuer, dass die Bio-Bauern nicht mehr konkurrieren k\"onnten. 
 -- Hinzu kommt ein Trend, der immer mehr Landwirte zu Energiewirten werden l\"asst: Nachwachsende Rohstoffe sind gefragt wie nie. \\ \hline
 \multirow{2}{\linewidth}{Consu- mer prices \& profit} & Acquisitions such as this takes away consumers' prerogative on where to spend our hard-earned dollars. 
 -- The consumer gains nothing from this. 
 -- But final cost to consumer is based on supply and demand. \\ \cline{2-2}
 & Beim Verbraucher bleibt so gut wie kein Preisvorteil. 
 -- Das man f\"ur "bessere" Erzeugnisse mehr zahlen muss, liegt auf der Hand.  
 -- Nur m\"ussen diese Billigartikel erst mal produziert werden, bevor der Verbraucher zugreifen kann. \\ \hline
\end{tabular}
\end{center}
\end{table*}

\end{document}